\documentclass[10pt,journal,compsoc]{IEEEtran}
\usepackage{amsmath}
\usepackage{amssymb}
\usepackage[english]{babel}
\usepackage{slashbox}
\usepackage{times}
\usepackage{epsfig}
\usepackage{graphicx}

\ifCLASSOPTIONcompsoc
  \usepackage[nocompress]{cite}
\else
  \usepackage{cite}
\fi
\ifCLASSINFOpdf
\else
\fi
\begin{document}
\title{Robust Non-Rigid Registration with Reweighted Position and Transformation Sparsity}

\author{Kun Li,~\IEEEmembership{Member,~IEEE,}~Jingyu Yang,~\IEEEmembership{Senior Member,~IEEE,}\\
        ~Yu-Kun Lai,~\IEEEmembership{Member,~IEEE,} and Daoliang Guo% <-this % stops a space

%E-mail: lik@tju.edu.cn
\IEEEcompsocitemizethanks{\IEEEcompsocthanksitem Jingyu Yang and Daoliang Guo are with the School of Electronic Information Engineering, Tianjin
University, Tianjin 300072, China.

\IEEEcompsocthanksitem Kun Li is with the Tianjin Key Laboratory of Cognitive Computing and
Application, School of Computer Science and Technology, Tianjin University,
Tianjin 300072, China.

\IEEEcompsocthanksitem Yu-Kun Lai is with the School of Computer Science and Informatics, Cardiff University, Wales, UK.\protect\\
Corresponding author: Kun Li (Email: lik@tju.edu.cn)}}


\markboth{Submitted to IEEE Transactions on Visualization and Computer Graphics}%
{Kun Li \MakeLowercase{\textit{et al.}}: Robust Non-Rigid Registration With Reweighted Dual Sparsities}



\IEEEtitleabstractindextext{%
\begin{abstract}
Non-rigid registration is challenging because it is ill-posed with high degrees of
freedom and is thus sensitive to noise and outliers.
We propose a robust non-rigid registration method using reweighted sparsities on position and transformation to estimate the deformations between 3-D shapes. We formulate the energy function with dual sparsities on both the data term and the smoothness term, and define the smoothness constraint using local rigidity.
The dual-sparsity based non-rigid registration model is enhanced with a reweighting scheme, and solved by
transferring the model into some alternating optimized subproblems which have exact solutions and guaranteed convergence. Experimental results on both public datasets and real scanned datasets show that our method outperforms the state-of-the-art methods and is more robust to noise and outliers than conventional non-rigid registration methods.

\end{abstract}

}

\maketitle

\IEEEdisplaynontitleabstractindextext

% \fi
\IEEEpeerreviewmaketitle

\IEEEraisesectionheading{\section{Introduction}\label{sec:introduction}}

%\begin{figure*}[ht]
%  \centering
%   \includegraphics[height=1.5in]{figure1}
%   \caption{ (a) template (b) target (c) result (d) fitting error (e)  target with outliers (f) result  (g) fitting error }
%  \label{fig:1}
%\end{figure*}
Non-rigid registration is a hot research topic in computer graphics and computer vision \cite{Tong2012Scanning, sidorov2011efficient, l0Norigid2015, dynamicfusion}, and is a key technique for dynamic 3-D reconstruction using a depth camera. Commodity depth sensors, e.g., Microsoft Kinect, become cheaper and more widely used, but depth images and reconstructed point clouds captured by such devices contain much noise. Hence, non-rigid registration methods robust to noise and outliers are highly desirable to scan dynamic scenes with deformable objects.

Given two input 3-D shapes, one as the template shape and the other as the target shape, non-rigid registration aims to find a suitable transformation that when applied deforms the template shape to be aligned with the target shape. Non-rigid registration is often formulated as an optimization problem. Most methods formulate some energy functional with both position and transformation constraints. The position constraint measures the closeness of the transformed template shape and the target shape, and the transformation constraint measures the fitness to model, which always includes the smoothness, namely the total energy of transformation differences of all the local neighbors. Most work uses the classic squared $\ell_2$-norm in the position constraint and the transformation constraint \cite{li2008global},\cite{amberg2007optimal},\cite{sorkine2007}.
However, the quadratic energy functional is more easily affected by noise and outliers. To address this problem,
Yang $et~al.$~\cite{ySNR2015} propose a sparse non-rigid
registration (SNR) method with an $\ell_1$-norm regularized model for the transformation constraint. However, their position constraint is still based on the $\ell_2$-norm. In practice, e.g. for near piece-wise rigid deformation, which is common for real-world deformable objects, the positional error tends to concentrate on small regions. This cannot be modeled well using the $\ell_2$-norm.

%Bouaziz $et~al.$\cite{bouaziz2013sparse}  propose a new  variant of the ICP algorithm  using sparsity-inducing norms to represent the data term.

%Based on these methods, we consider to constrain the data term as well as smooth term in $\ell_1$-norm.  $\ell_1$-norm can be viewed as a selection of important features, it can make the function sparse. The transform and positional distance of two 3D shapes always smooth in most area, except only  a small part (e.g.,joint ). $\ell_1$-norm can represent this feature very well.

In this paper, we propose a non-rigid registration method with sparsity-regularized position and transformation constraints.
The distribution of positional errors and transformation differences for typical non-rigid deformation can be well modeled using the Laplacian distribution, or equivalently, the $\ell_1$-norm should be used to measure both the positional errors and transformation differences, which is therefore called \emph{dual sparsities}.
To promote the sparsity, we adopt a reweighted sparse model, which is solved by the alternating direction method of multipliers (ADMM). The proposed method is evaluated on public datasets~{\cite{bronstein2008numerical,vlasic2008articulated}} and real datasets captured by a RGB-D depth sensor. The results demonstrate that the proposed method obtains better results than the state-of-the-art non-rigid registration methods.

The main contributions of this work are summarized as:
\begin{itemize}
\item We propose a dual-sparsity based non-rigid registration method on both position and transformation constraints. The proposed model is robust against outliers as the sparsity terms allow a small fraction of regions with larger deviations.
\item We incorporate orthogonality  constraints in the dual-sparsity non-rigid registration framework to promote locally rigid transformations.
\item We equip the dual-sparsity based non-rigid registration model with a reweighted scheme to iteratively enhance sparsity in the series of alternating optimization subproblems.
\end{itemize}

\section{Related Work}
3-D shape registration consists of rigid registration and non-rigid registration.
Rigid registration aims to find a \emph{global} rigid-body transformation, while non-rigid registration needs to find a set of \emph{local} transformations that align two shapes.

\begin{figure*}[!ht]
\centering
\includegraphics[width =
6in]{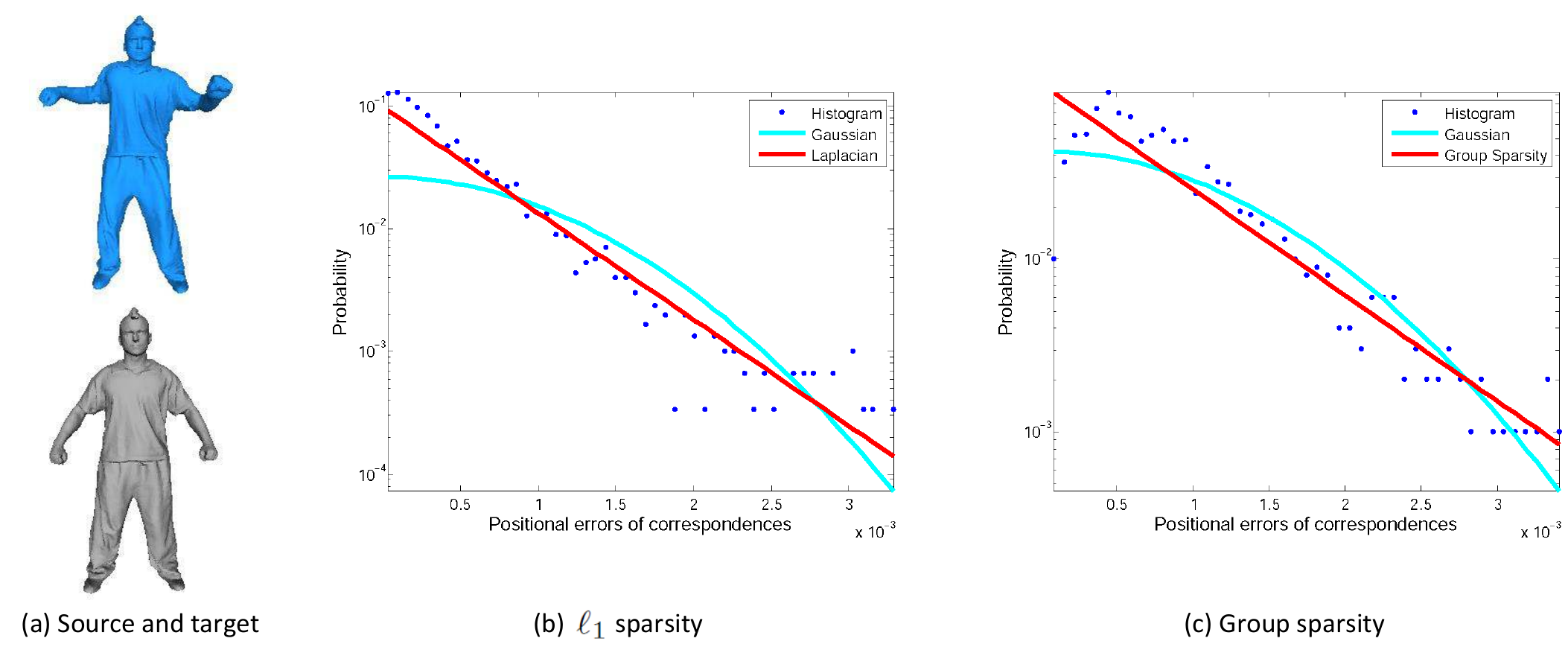}
\caption{Normalized histograms and the associated fitted Laplacian and Gaussian
distributions of positional errors measured in the $\ell_1$ norm (with equal contribution from each dimension) (b) or with Euclidean distance (c) for \emph{Bouncing} dataset (a). }
\label{fig:onecol}
\end{figure*}

In rigid registration, the 3-D shapes are assumed to be aligned by a Euclidean transformation, including rotation and
translation. Iterative Closest Point (ICP) and its variants \cite{besl1992method} are the dominant algorithms for rigid registration. This kind of methods alternates between two steps: 1) finding closest points and 2) solving the optimal transformation. As an improved method of ICP, Chen \emph{et al.}~\cite{chen1991object} minimize the shortest
distance between a point in the template and the tangent plane of the
closest point on the target. Pottmann \emph{et al.}~\cite{pottmann2004registration} propose a registration
method with quadratic convergence, which gives faster and more stable convergence than the standard ICP ~\cite{pottmann2006geometry}.
Bouaziz $et~al.$ \cite{bouaziz2013sparse}  propose a new  variant of the ICP algorithm, which uses sparsity-inducing norms to represent the positional constraint, and achieve better results for the situation with noise and outliers. Their work focuses on rigid registration with low degrees of freedom, and hence regularization is not necessary.

%Flory \emph{et al.}\cite{Flory:2010:SFR} propose  a technique for local, rigid registration of surfaces or point clouds fitting to point sets based on the $\ell_1$-norm.  Their technique performs better than $\ell_2$-methods in the presence of outliers.

When shapes have large deformations from template to target, automatic non-rigid registration is necessary.
It is more challenging due to its high degrees of freedom, and an appropriate deformation model is the key for an efficient and robust algorithm.

Some methods compute global rigid transformations for bones and local non-rigid transformations near joints, which is essentially a piecewise rigid transformation model.
Allen \emph{et al.}~\cite{allen2002articulated} place markers on the object to help reconstruct the pose of scan and use it  as a basis for modeling deformation. Pekelny \emph{et al.}~\cite{pekelny2008articulated} use predefined bone information to find  bone transformations.

Some models take more generic deformations into consideration.
Chui \emph{et al.}~\cite{chui2003new} use the thin-plate spline (TPS) as the non-rigid transformation model.
 Papazov \emph{et al.}~\cite{papazov2011deformable} allow points to move freely and use an additional uniform distribution to limit noise and outliers, and propose an ordinary differential equation (ODE) model.
Local affine transformations~\cite{allen2003space} are also frequently used in non-rigid registration.
Liao \emph{et al.}~\cite{liao2009modeling} use differential coordinates as local affine transformations with smoothness constraints. Amberg \emph{et al.}~\cite{amberg2007optimal} use a  stiffness term to ensure similarity of adjacent transformations. Rouhani \emph{et al.}~\cite{rouhani2014non} model non-rigid deformation as an integration of locally rigid transformations. In our work, we use local affine transformation with an orthogonality constraint as it allows more flexibility to capture fine surface details while keeping local shapes.

Non-rigid registration is often formulated as an energy functional with data and regularization terms.
Most of the non-rigid registration work models the data term in $\ell_2$-norm in a least-squares sense~\cite{sumner2004deformation},\cite{amberg2007optimal}.

Regularization terms help to preserve smoothness, making the optimization more robust to noise and outliers,
and $\ell_2$-norm is also widely used in regularization terms.
S\"{u}{\ss}muth  \emph{et al.} \cite{sussmuth2008reconstructing}  use a generalized as-rigid-as-possible energy~\cite{sorkine2007} to promote smoothness. Liao \emph{et al.}~\cite{liao2009modeling} define a transformation
model using the TPS~\cite{chui2003new}, and use graduated assignment for non-rigid registration and optimization. Wand \emph{et al.}~\cite{wand2009efficient} take a set of time-varying point data as input, and reconstruct a single shape and a deformation field that fit the data. To improve robustness, Li \emph{et al.}~\cite{li2008global} solve correspondences, confidence weights, and a deformation field within a single optimization framework using $\ell_2$-norm. Hontani \emph{et al.}~\cite{Hontani2012outliersparsity} propose a statistical shape model (SSM) which is incorporated into the nonrigid ICP (NICP), and outliers can be detected based on their sparseness. Yang $et~al.$~\cite{ySNR2015} propose a sparse non-rigid registration (SNR) method with an $\ell_1$-norm regularized model for the smoothness. However, their $\ell_2$-norm position constraint cannot model the concentricity of positional errors well.

In this paper, based on the observation that the deformations of 3-D surfaces vary smoothly and the positional distances and transformation differences are sparse, we propose a non-rigid registration method with sparse position and transformation constraints. The model is efficiently solved by the alternating direction method under the augmented Lagrangian multiplier framework.

\section{Motivation}
\label{sec:motivation}

The traditional quadratic data term assumes the Gaussian distribution of positional errors. However, transformations are piecewise smooth signals residing on 3D surfaces, resulting in larger positional errors for geometric details and joints. This suggests that the positional errors are sparse, and should be modeled by a heavy-tailed distribution, rather than being dense and modeled by a rapidly vanishing Gaussian distribution. This is verified in Fig. \ref{fig:onecol}(b). We uniformly pick up $10\%$ ground truth matchings (vertices) as correspondences, and solve the transformations using the SNR method \cite{ySNR2015} which measures the positional errors in the standard quadratic term to avoid bias towards the $\ell_1$-norm. The Laplacian distribution fits the histogram significantly better than the Gaussian distribution, suggesting the use of sparsity-promoting $\ell_1$-norm in the data term.

Our $\ell_1$-norm sparsity measures equally coordinate differences of each dimension. Another possibility is to use the sum of Euclidean distances (group sparsity) between corresponding points, which also well fits the distribution of positional errors as shown in Fig. \ref{fig:onecol}(c). The group sparsity advocates sparsity for each Euclidean distance as a whole, while the $\ell_1$-norm allows a large distance along a particular dimension although the Euclidean distance is not significant. In this sense, $\ell_1$-norm is more flexible to preserve large non-rigid deformation along some dimensions.
Such an advantage is also observed in the anisotropic total variation (TV) \cite{decomposition2004} that applies the $\ell_1$-norm on the image gradient over the isotropic TV \cite{rudin1992nonlinear} that measures TV as the sum of $\ell_2$-norm (not squared). Birkholz \cite{Birkholz20112502} showed that anisotropic TV achieves better denoising performance in preserving the geometries of corners in images. We choose the $\ell_1$-norm to measure the positional errors for its potential flexibility, and also for its easier and faster implementation with an element-wise shrinkage (cf. Table 1 for statistics of running times).

\section{The Proposed Method}

\label{sec:model}
\subsection{Iterative Framework}
\begin{figure*}[ht]
  \centering
  \includegraphics[width=6.0in]{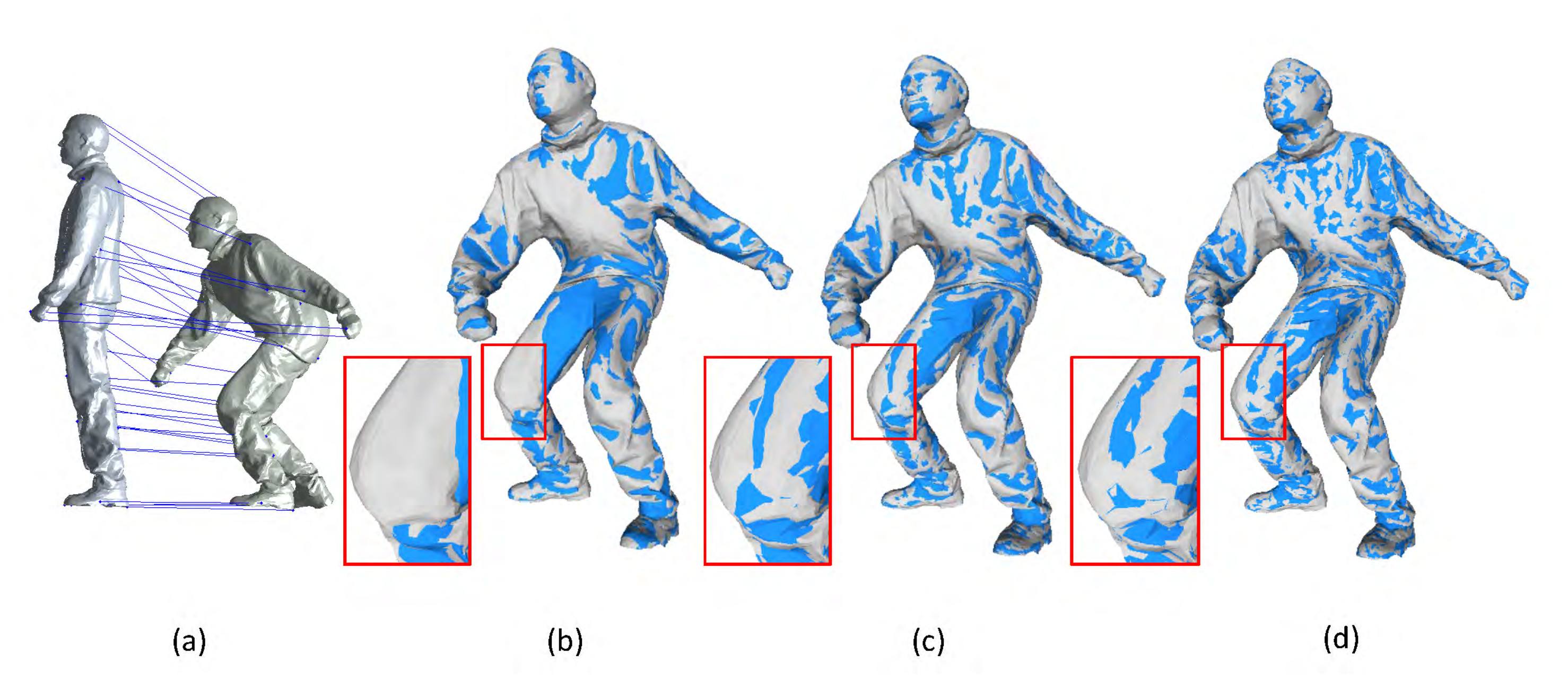}
  \caption{Comparison results on \emph{Jumping} dataset with 35
manually-specified correspondences: (a) Given correspondences,
(b) $\ell_2$-norm method, (c) SNR method \cite{ySNR2015}, and (d) Our method}
  \label{fig:few_corr}
\end{figure*}

We iteratively compute the deformation between the template shape and the target shape. Each iteration consists of two steps. In the first step,  the correspondences between template and target are estimated using the registration result obtained from the last iteration. At the beginning of the iteration, we use a  technique based on local geometric similarity and diffusion pruning of inconsistent correspondence~\cite{tam2014diffusion} as it often provides reliable correspondences. Alternative
correspondence techniques or manual specification of a few
correspondences may instead be used (an example is shown in  Fig. \ref{fig:few_corr}). These computed correspondences are referred to as the correspondence mapping $f$.
 Then, we use the closest points between template and target shapes to find additional correspondences similar to ICP.
 In the second step (Sec. \ref{sec:transformation_estimation}),
 we propose an energy-minimization approach based on dual-sparsity representation to estimate  the non-rigid transformations using the correspondences obtained from the first step.

 \subsection{Deformation Estimation}
\label{sec:transformation_estimation}
\noindent
Let  $\mathbf{v}_i \triangleq [x_i, y_i, z_i, 1]^{\top}$ be a 3D point in the homogenous coordinate.
 Denote by $\mathcal{V}\triangleq\{ {\mathbf{v}_1}, \cdots, \mathbf{v}_{N} \}$ a template set of 3D points and by $\mathcal{U}\triangleq\{ \mathbf{u}_1, \cdots, \mathbf{u}_{M} \}$ a target set of 3D points, where $N$ and $M$ are the numbers of points.
Denote by $\mathbf{u}_{f(i)} \in \mathcal{U}$ the correspondence of $\mathbf{v}_{i} \in \mathcal{V}$. Define
$f:\{1, \cdots, N\} \mapsto \{0, 1, \cdots, M\}$ as the index mapping from the template points to the target points, where $f(i) = 0$ means the corresponding vertex cannot be found for the $i$-th vertex.
Denote by  $\mathbf{X}_i$ the $3\times 4$ transformation matrix for point $\mathbf{v}_i$. Define $\mathcal{X}\triangleq \left\{\mathbf{X}_1, \cdots, \mathbf{X}_{N}\right\}$ as the set of non-rigid transformations.
For compact notation, we define $\mathbf{X} \triangleq \left[ \mathbf X_1, \cdots, \mathbf X_N \right]^{\top}$ as a matrix containing the $N$ transformation matrices to be solved. The proposed method is to find non-rigid transformations $\mathbf X$ that transforms the template $\mathcal V$ into the target  $\mathcal U$ as accurately as possible, given a correspondence mapping $f$.

The non-rigid registration is formulated as the minimization of the following energy function:
\begin{equation}\label{equ:E_sum}
 E\left(\mathbf{X}; f\right)= E_{data}\left( \mathbf{X}; f \right) + \alpha E_{smooth}\left( \mathbf{X} \right) + \beta E_{orth}\left( \mathbf{X} \right),
\end{equation}
 where $E_{data}\left( \mathbf{X} \right)$, $E_{smooth}\left( \mathbf{X} \right)$ and $E_{orth}\left( \mathbf{X} \right)$  are  data term, smoothness term, and orthogonality constraint, respectively. $\alpha$ and $\beta$ adjust the importance of different terms. The data term measures the position accuracy, the smoothness term imposes a smoothness constraint so that the original ill-posed problem (defined by only the data term) is now well-posed, and the orthogonality constraint promotes locally rigid transformations, which is particularly needed for underconstrained scenarios such as partial meshes.

 \noindent\textbf{\textit{Data term:}}
We measure the accuracy of deformation as the closeness of the transformed points to their corresponding target points.
We assign a weight, denoted by $w_i$, for each point. The weight $w_i$ is one if there is a corresponding point on the target shape for $\mathbf{v}_i$, and zero otherwise. Hence, we propose the following data term
\begin{equation}\label{equ:E_data_sum}
\begin{split}
E_{data}\left( \mathbf{X}; f\right) &= \sum_{\mathbf v_i\in \mathcal V} {w_i\begin{Vmatrix}\mathbf X_i\mathbf v_i- \tilde{\mathbf{u}}_{f(i)} \end{Vmatrix}}_1,\\
\end{split}
\end{equation}
where $\tilde{\mathbf{u}}_{f(i)}$ is the Cartesian coordinate of $\mathbf{u}_{f(i)}$.

For the compact representation in algorithm derivation, we define the following matrix/vector form of the variables to reformulate data term (\ref{equ:E_data_sum}):
\begin{eqnarray}
\begin{split}
\mathbf{W} &= \textrm{diag} \left(\sqrt{w_1}, \cdots, \sqrt{w_N} \right),\\
\mathbf{V} &= \textrm{diag}\left( \mathbf{v}_1^{\top}, \cdots, \mathbf{v}_N^{\top} \right),\\
\tilde{\mathbf{U}}_f &= \begin{bmatrix} \tilde{\mathbf{u}}_{f(1)}&\cdots &\tilde{\mathbf{u}}_{f(N)}\end{bmatrix}^{\top},
\end{split}
\end{eqnarray}
where $\textrm{diag}(\cdot)$ is a diagonal matrix containing the input elements as diagonal entities. Then, the data term can be rewritten as
\begin{equation}\label{equ:E_data_matrix}
E_{data}\left( \mathbf{X}; f\right) = \begin{Vmatrix}\mathbf W  \left( \mathbf V\mathbf X-\tilde{\mathbf{U}}_f \right) \end{Vmatrix}_1.
\end{equation}

\noindent\textbf{\textit{Smoothness term:}}
In the smoothness term, local rigidity is assumed: for vertex $\mathbf v_i$, the transformations of neighboring vertices $\mathbf v_j\in \mathcal{N}_i$ should have very close transformed positions when applied to $\mathbf v_i$. Therefore, we define the following  smoothness term:
\begin{equation}
\begin{split}
E_{smooth}\left( \mathbf{X} \right) &= \sum_{\mathbf v_i\in \mathcal V}\sum_{\mathbf v_j\in \mathcal{N}_i} {\begin{Vmatrix} \mathbf X_i \mathbf  v_i- \mathbf X_j \mathbf v_i \end{Vmatrix}}_1.
\end{split}
\label{equ:E_smooth_sum_new}
\end{equation}

Define a graph $\mathcal{G} \triangleq \left( \mathcal{V},  \mathcal{E}\right)$, where the vertices of the graph are the 3D points in $\mathcal{V}$, and the edges of the graph are denoted by $\mathcal{E}$. For a 3D mesh, edges of the graph are simply defined by the edges of the mesh; for 3D point clouds, edges can be defined by {connecting each vertex with its $K$-nearest neighbors ($K$ is typically set to 6).} Denote the neighborhood of vertex $\mathbf{v}_i$ by $\mathcal{N}_i$, and an edge $e_{ij}$ is defined between each neighboring vertex $\mathbf{v}_j$ and $\mathbf{v}_i$. So, we have $\mathcal{E} = \left\{e_{ij} \mid \mathbf{v}_j\in \mathcal{N}_i, \mathbf{v}_i \in \mathcal{V}\right\}$. Similar to the data term, we define a differential matrix $\mathbf K \in \{-1, 1\}^{|\mathcal{E}|\times |\mathcal{V}|}$ on the graph $\mathcal{G}$ for concise presentation. Concretely, each row of $ \mathbf K $ corresponds to an edge in $\mathcal{E}$ and each column corresponds to a vertex in $\mathcal{V}$. Each row in $\mathbf{K}$ has only two nonzero entries. For example, assuming the $r^{\textrm{th}}$ row in $\mathbf{K}$ associates with edge $e_{ij}$, then the entry related to the reference vertex $\mathbf{v}_i$ is set at 1, while the one related to the neighboring vertex $v_j$ is set at -1, i.e. $k_{ri} = 1$ and $k_{rj} = -1$.
Let $\mathbf k_{i:}$ denote the $i^\textrm{th}$ row of $\mathbf K$. We introduce a matrix $\mathbf B \in R^{|\mathcal{E}|\times 4|\mathcal{V}|}$, where the $i^{\textrm{th}}$ row of $\mathbf B$ is defined as $\mathbf b_{i:} := \mathbf k_{i:}\otimes \mathbf v_i^\top$. Therefore, the cost of transformation smoothness is rewritten as
\begin{equation}\label{equ:E_smooth_matrix}
   E_{smooth}\left( \mathbf{X} \right) =  \begin{Vmatrix} \mathbf B \mathbf X \end{Vmatrix}_1.
\end{equation}

\noindent\textbf{\textit{Orthogonality constraint:}}
Especially for partial meshes with large motions, the problem may be underconstrained leading to large distortions. In this case, orthogonality constraint is effective in better preserving local shapes and making the solution more reasonable.
\begin{equation}
\begin{split}
&E_{orth}\left( \mathbf{X} \right) = \sum_{i=1}^N {\begin{Vmatrix} \mathbf S_i \mathbf X_i - \mathbf R_i \end{Vmatrix}}_F^2,\\
&~~\textrm{s.t.}\quad \mathbf R_i^T \mathbf R_i = \mathbf I, ~det(\mathbf R_i) > 0,
\end{split}
\label{equ:E_smooth_sum_new}
\end{equation}
where $\mathbf R_i$ is a $3 \times 3$ rotation matrix, and $\mathbf S_i$ is a constant $3 \times 4$ matrix that extracts
the rotation component of $\mathbf X_i$. $det(\mathbf R_i)>0$ ensures that $\mathbf R_i$ is a rotation matrix, not a mirrored matrix.

The final energy function has the following compact form with matrix-vector notations:
 \begin{equation}
 \label{equ:E_matrix_new}
 \begin{split}
 &\min_{\mathbf X} \begin{Vmatrix}\mathbf W \left( \mathbf V\mathbf X-\tilde{\mathbf{U}}_f \right) \end{Vmatrix}_1 + \alpha \begin{Vmatrix}  \mathbf B \mathbf X\end{Vmatrix}_1 + \beta\sum_{i=1}^N {\begin{Vmatrix} \mathbf S_i \mathbf X_i - \mathbf R_i \end{Vmatrix}}_F^2, \\
&\quad \quad \quad \quad \quad \quad \textrm{s.t.}\quad \mathbf R_i^T \mathbf R_i = \mathbf I, ~det(\mathbf R_i) > 0.
\end{split}
 \end{equation}
 \noindent\textbf{\textit{Reweighting:}}
 To further promote sparsity, both the data term and the smoothness term are weighted, and the weighting matrices are updated at each iteration of non-rigid registration.  The weighted version of the dual-sparsity model (\ref{equ:E_matrix_new}) is defined as follows:
 \begin{equation}
 \label{equ:E_matrix_weighted}
 \begin{split}
 &\min_{\mathbf X} \begin{Vmatrix}\mathbf W_{\textrm{D}}\left( \mathbf V\mathbf X-\tilde{\mathbf{U}}_f \right) \end{Vmatrix}_1 + \alpha \begin{Vmatrix}  \mathbf W_{\textrm{S}} \mathbf B \mathbf X\end{Vmatrix}_1 + \beta\sum_{i=1}^N {\begin{Vmatrix} \mathbf S_i \mathbf X_i - \mathbf R_i \end{Vmatrix}}_F^2, \\
&\quad \quad \quad \quad \quad \quad  \textrm{s.t.}\quad \mathbf R_i^T \mathbf R_i = \mathbf I, ~det(\mathbf R_i) > 0.
\end{split}
 \end{equation}
where $\mathbf W_{\textrm{D}}$ and $\mathbf W_{\textrm{S}}$ are diagonal weighting matrices for the data term and smoothness term, respectively. The weighting matrices are updated according to the $\ell_1$-norm of the corresponding entries. For the data term, the weights are updated as
 \begin{equation}
 \label{equ:weighting_dataterm}
\mathbf W^{(l)}_{\textrm{D}}(i,i) =
\begin{cases}
\frac{1}{{\left\|\mathbf X_i^{(l-1)}\mathbf v_i- \tilde{\mathbf{u}}_{f(i)}^{(l)} \right\|}_1 + \epsilon_{\textrm{D}}}, & f(i) \neq 0,\\
0, & f(i) = 0,
\end{cases}
\end{equation}
where $l$ represents the index of iteration, $\epsilon_{\textrm{D}}$ is a constant to avoid the division-by-zero issue, and is set as 0.01 in the experiments.
Similarly, the weights for the smoothness term are updated as
 \begin{equation}
 \label{equ:weighting_smoothterm}
\mathbf W^{(l)}_{\textrm{S}}(r,r) =
\frac{1}{{\left\| \mathbf X_i^{(l-1)} \mathbf  v_i- \mathbf X^{(l-1)}_j \mathbf v_i \right\|}_1 + \epsilon_{\textrm{S}}},
\end{equation}
where $\epsilon_{\textrm{S}}$ is a constant which is set as 0.01 in the experiments, and the $r^{\textrm{th}}$ row of matrix $\mathbf B \mathbf X$ is associated with edge $e_{ij}$ between $\mathbf v_i$ and  $\mathbf v_j$.

%Regarding the minimization (\ref{equ:E_matrix_weighted}) given weighting matrices $\mathbf W_{\textrm{D}}$ $\mathbf W_{\textrm{S}}$, the algorithm  can be readily derived in the same way as minimization (\ref{equ:E_matrix_new}).

To solve the problem, we first transform the minimization (\ref{equ:E_matrix_weighted}) into the following form with auxiliary variables $\mathbf A$ and $\mathbf C$:
 \begin{equation}
 \label{equ:E_matrix_constrained_WS_WD}
 \begin{split}
 \min_{\mathbf X, \mathbf C, \mathbf A} &\begin{Vmatrix} \mathbf C \end{Vmatrix}_1 + \alpha \begin{Vmatrix} \mathbf A\end{Vmatrix}_1 + \beta\sum_{i=1}^N {\begin{Vmatrix} \mathbf S_i \mathbf X_i - \mathbf R_i \end{Vmatrix}}_F^2,\\
 &\textrm{s.t.}\quad \mathbf C = \mathbf W_{\textrm{D}} \left( \mathbf V\mathbf X-\tilde{\mathbf{U}}_f \right),\\
& \quad \quad\mathbf A = \mathbf W_{\textrm{S}} \mathbf B \mathbf X, \\
&\quad \quad \mathbf R_i^T \mathbf R_i = \mathbf I, ~det(\mathbf R_i) > 0.
\end{split}
 \end{equation}
Then, we solve the constrained minimization (\ref{equ:E_matrix_constrained_WS_WD}) using the augmented Lagrangian method (ALM) \cite{Bertsekas82}. The ALM method converts the original problem (\ref{equ:E_matrix_constrained_WS_WD}) to iterative minimization of its augmented Lagrangian function:
 \begin{equation}\label{augmented_Lagrangian_WS_WD}
\begin{split}
&L(\mathbf X, \mathbf C, \mathbf A, \{\mathbf R_i\}, \mathbf Y_1, \mathbf Y_2, \mu_1, \mu_2) = \begin{Vmatrix}\mathbf C \end{Vmatrix}_1 \ + \alpha \begin{Vmatrix}\mathbf A\end{Vmatrix}_1 \\
&+ \left\langle \mathbf Y_1, \mathbf C - \mathbf W_{\textrm{D}} \left( \mathbf V\mathbf X-\tilde{\mathbf{U}}_f \right) \right\rangle\\
&+ \frac{\mu_1}{2}\begin{Vmatrix}\mathbf C - \mathbf W_{\textrm{D}} \left( \mathbf V\mathbf X-\tilde{\mathbf{U}}_f \right) \end{Vmatrix}_F^2\\
&+ \langle \mathbf Y_2,\mathbf A-\mathbf W_{\textrm{S}}\mathbf B\mathbf X\rangle +\frac{\mu_2}{2}\begin{Vmatrix}\mathbf A -\mathbf W_{\textrm{S}}\mathbf B\mathbf X\end{Vmatrix}_F^2 \\
&+ \beta\sum_{i=1}^N {\begin{Vmatrix} \mathbf S_i \mathbf X_i - \mathbf R_i \end{Vmatrix}}_F^2,\\
&\textrm{s.t.}\quad \mathbf R_i^T \mathbf R_i = \mathbf I, ~det(\mathbf R_i) > 0,
\end{split}
 \end{equation}
where ($\mu_1$, $\mu_2$) are positive constants, ($\mathbf Y_1$,$\mathbf Y_2$) are Lagrangian multipliers, and $\langle \cdot , \cdot \rangle $ denotes the inner product of two matrices considered as long vectors. Under the standard ALM framework, ($\mathbf Y_1$, $\mathbf Y_2$ ) and ($\mu_1$, $\mu_2$) can be efficiently updated. However, each iteration has to solve  $\mathbf A$, $\mathbf C$, $\{\mathbf R_i\}$ and $\mathbf X$ simultaneously, which is difficult yet computationally demanding. Hence, we resort to the alternate direction method (ADM) \cite{Boyd11} to optimize  $\mathbf A$, $\mathbf C$, $\{\mathbf R_i\}$ and $\mathbf X$ separately at each iteration:
\begin{equation}\label{equ:solveCAX_WS_WD}
\begin{cases}
\mathbf C^{(k+1)} = &\arg\min_{\mathbf C}~ \|\mathbf C\|_1\\
&+\left\langle \mathbf Y_1^{(k)}, \mathbf C - \mathbf W_{\textrm{D}} \left( \mathbf V\mathbf X^{(k)}-\tilde{\mathbf{U}}_f \right) \right\rangle\\
&+\frac{\mu_1^{(k)}}{2}\begin{Vmatrix}\mathbf C - \mathbf W_{\textrm{D}} \left( \mathbf V\mathbf X^{(k)}-\tilde{\mathbf{U}}_f \right) \end{Vmatrix}_F^2,\\
\mathbf A^{(k+1)}=&\arg\min_{\mathbf A}~\alpha \|\mathbf A\|_1+\left\langle \mathbf Y_2^{(k)},\mathbf A-\mathbf W_{\textrm{S}}\mathbf B\mathbf X^k\right\rangle\\
& +\frac{\mu^{(k)}_2}{2}\begin{Vmatrix}\mathbf A-\mathbf W_{\textrm{S}}\mathbf B\mathbf X^{(k)}\end{Vmatrix}_F^2,\\
\mathbf R_i^{(k+1)}=& \arg\min_{\mathbf R_i}~ \beta\sum_{i=1}^N {\begin{Vmatrix} \mathbf S_i \mathbf X_i^{(k)} - \mathbf R_i \end{Vmatrix}}_F^2\\
& \textrm{s.t.}\quad \mathbf R_i^T \mathbf R_i = \mathbf I,  ~det(\mathbf R_i) > 0\\
\mathbf X^{(k+1)}=& \arg\min_{\mathbf X}~ \left\langle \mathbf Y_1^{(k)}, \mathbf C^{(k+1)} - \mathbf W_{\textrm{D}} \left( \mathbf V\mathbf X-\tilde{\mathbf{U}}_f \right) \right\rangle \\
&+\frac{\mu_1^{(k)}}{2}\begin{Vmatrix}\mathbf C^{(k+1)} - \mathbf W_{\textrm{D}} \left( \mathbf V\mathbf X-\tilde{\mathbf{U}}_f \right) \end{Vmatrix}_F^2\\
&+\left\langle \mathbf Y_2^{(k)}, \mathbf A^{(k+1)}-\mathbf W_{\textrm{S}}\mathbf B\mathbf X\right\rangle \\ &+\frac{\mu_2^{(k)}}{2}\begin{Vmatrix}\mathbf A^{(k+1)}-\mathbf W_{\textrm{S}}\mathbf B\mathbf X\end{Vmatrix}_F^2\\
&+ \beta\sum_{i=1}^N {\begin{Vmatrix} \mathbf S_i \mathbf X_i - \mathbf R_i^{(k+1)} \end{Vmatrix}}_F^2,\\
\mathbf Y_1^{(k+1)}=&\mathbf Y_1^{(k)} + \mu_1^{(k)} \left(\mathbf C^{(k+1)} - \mathbf W_{\textrm{D}} \left( \mathbf V\mathbf X^{(k+1)}-\tilde{\mathbf{U}}_f \right)\right),\\
\mathbf Y_2^{(k+1)}=&\mathbf Y_2^{(k)} + \mu_2^{(k)} \left(\mathbf A^{(k+1)}-\mathbf W_{\textrm{S}}\mathbf B\mathbf X^{(k+1)}\right),\\
~\mu_1^{(k+1)}=&\rho_1\mu_1^{(k)}, ~~\rho_1 > 1,\\
~\mu_2^{(k+1)}=&\rho_2\mu_2^{(k)}, ~~\rho_2 > 1.
\end{cases}
\end{equation}

The $\mathbf{C}$-subproblem has the following closed solution:
\begin{equation}\label{equ:C_solution_WS_WD}
\begin{split}
 & \mathbf C^{(k+1)} =\\
  &\textrm{shrink}\left(\mathbf W_{\textrm{D}} \left( \mathbf V\mathbf X^{(k)}-\tilde{\mathbf{U}}_f \right)-\frac{1}{\mu_1^{(k)}}\mathbf Y_1^{(k)},\frac{1}{\mu_1^{(k)}}\right),
 \end{split}
 \end{equation}
 where shrink($\cdot$,$\cdot$) is the shrinkage function applied on the matrix element-wise:
\begin{equation}\label{equ:shinkage_WS_WD}
\textrm{shrink}\left(x, \tau\right) = \textrm{sign}(x)\max(|x|-\tau, 0).
 \end{equation}

 The $\mathbf{A}$-subproblem is solved in a similar way:
\begin{equation}\label{equ:A_solution_WS_WD}
 \mathbf A^{(k+1)} = \textrm{shrink}\left(\mathbf W_{\textrm{S}}\mathbf B\mathbf X^{(k)}-\frac{1}{\mu_2^{(k)}}\mathbf Y_2^{(k)},\frac{\alpha}{\mu_2^{(k)}}\right).
 \end{equation}

 The $\mathbf{R}_i$-subproblem can be explicitly solved using Procrustes projection:
\begin{equation}\label{equ:R_solution_WS_WD}
\begin{split}
 \mathbf U \mathbf D \mathbf V^T &= \textrm{svd}(\mathbf S_i \mathbf X_i^{k}),\\
 \mathbf R_i^{k+1} &= \mathbf U \mathbf \mathbf V^T.
\end{split}
 \end{equation}
If the obtained matrix has a negative determinant, take $\mathbf R_i$ with the opposite sign to turn the matrix into a rotation matrix.

 \begin{figure*}[!t]
  \centering
  \includegraphics[width=0.95\linewidth]{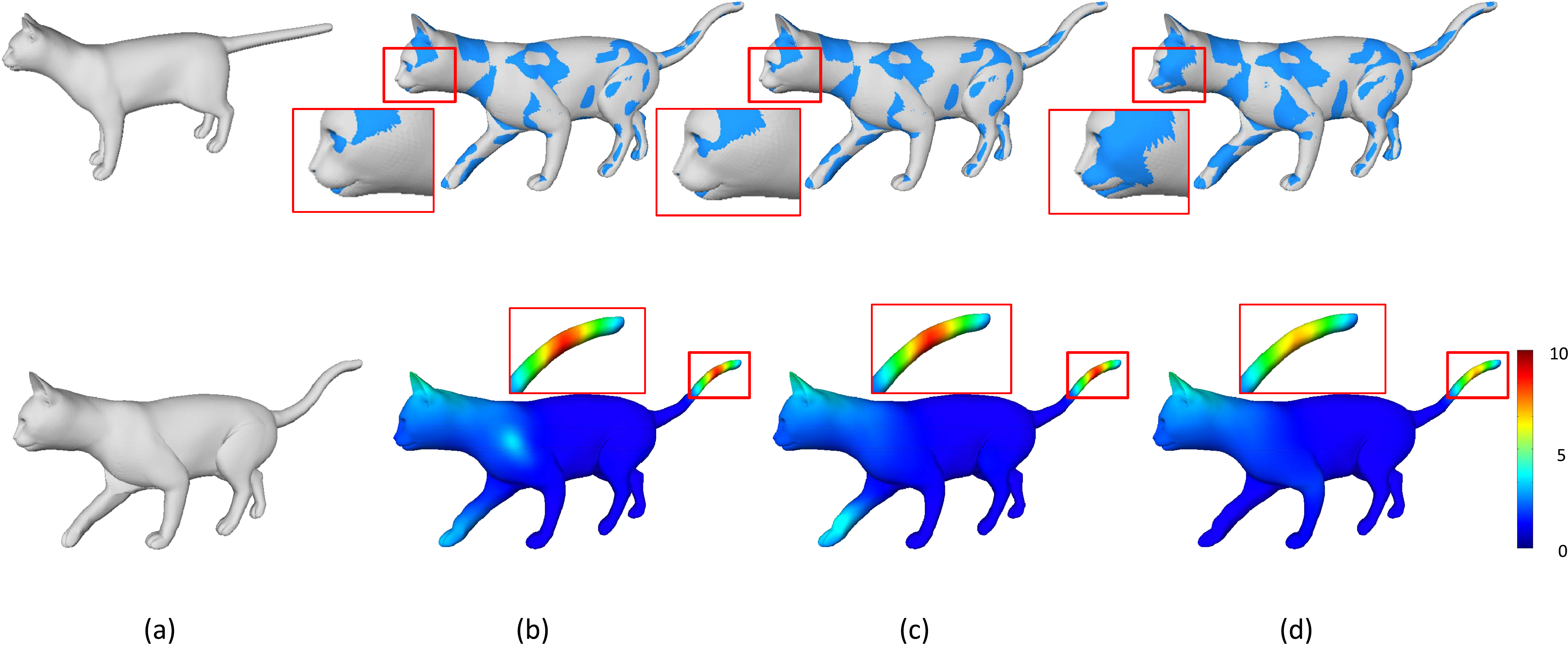}
  \caption{(a) Template (top) and target (bottom) shapes, (b)-(d): Comparison results (top) and fitting errors (bottom)  of (b) $\ell_2$-norm method, (c) SNR method \cite{ySNR2015} and (d) Our method on \emph{Cat} dataset.}
  \label{fig:L2_L2L1_L1_cat}
\end{figure*}

\begin{figure*}[!t]
  \centering
  \includegraphics[width=0.95\linewidth]{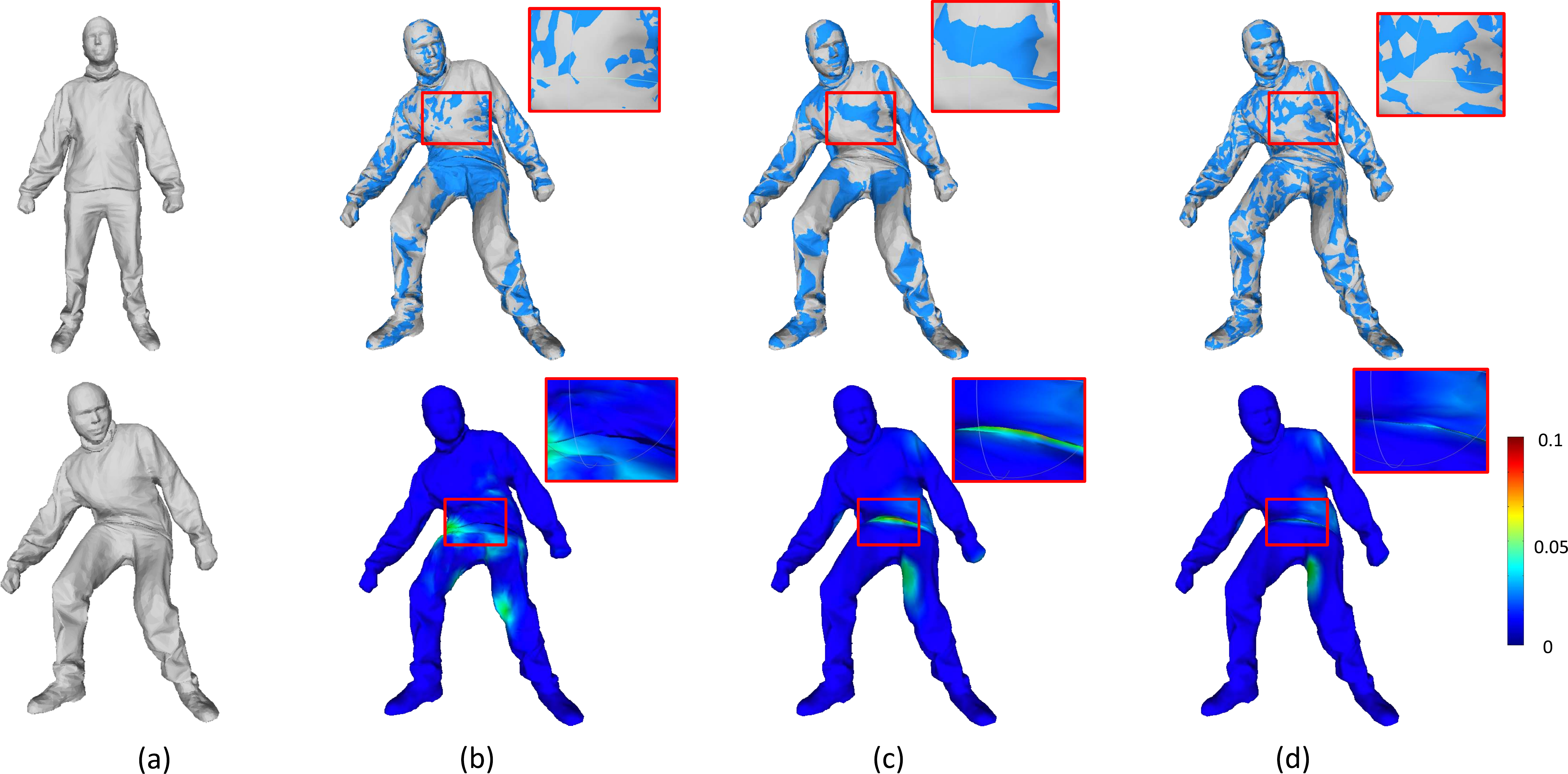}
  \caption{(a) Template (top) and target (bottom) shapes, (b)-(d): Comparison results (top) and fitting errors (bottom)  of (b) $\ell_2$-norm method, (c) SNR method \cite{ySNR2015} and (d) Our method on \emph{Jumping} dataset.}
  \label{fig:L2_L2L1_L1_human}
\end{figure*}

 \begin{figure*}[ht]
  \centering
  \includegraphics[width=6.0in]{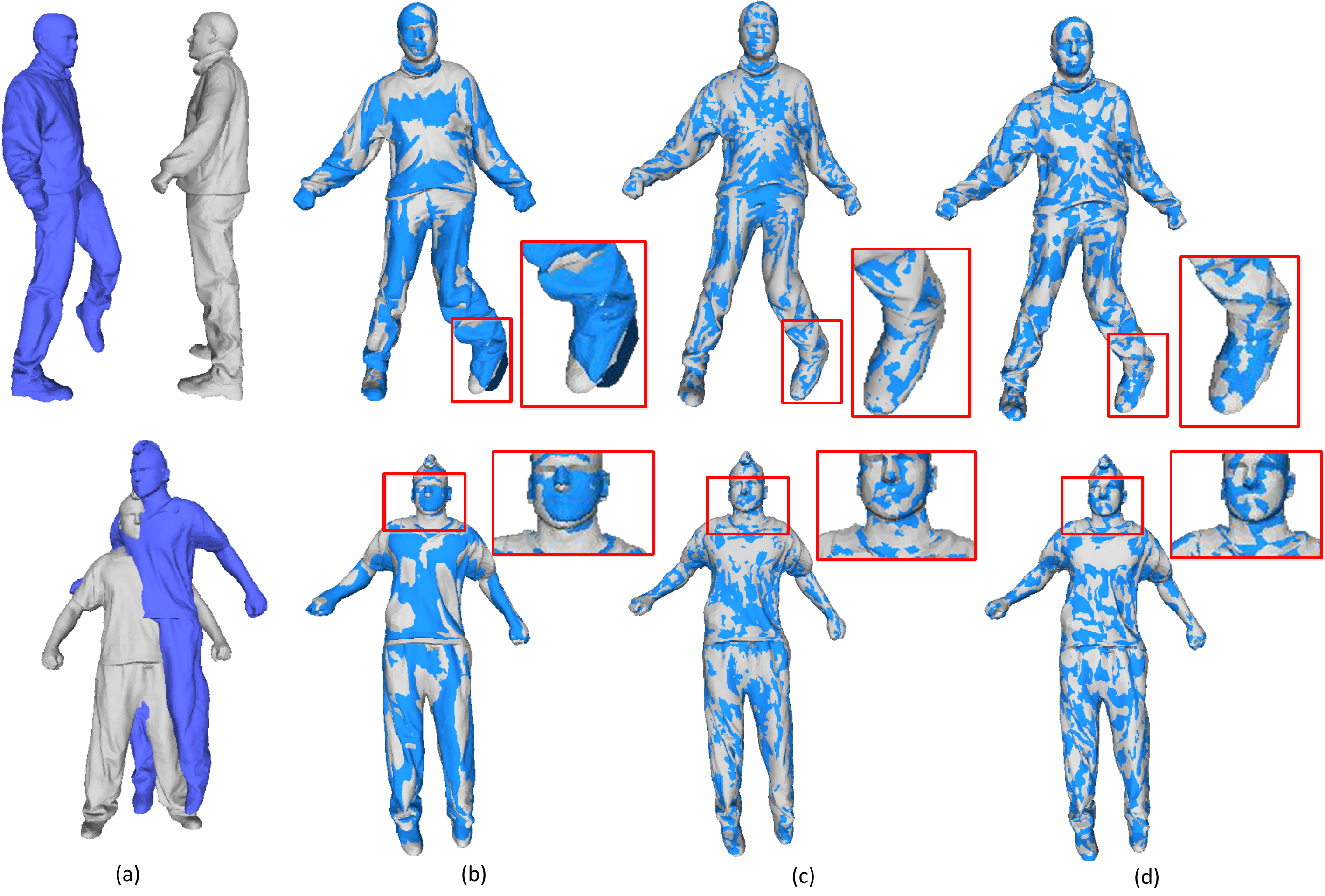}
  \caption{Comparison results on \emph{Bouncing} dataset: (a)
Template and target, (b) The method in \cite{li2008global}, (c) SNR method \cite{ySNR2015}, and (d) Our
method.}
  \label{fig:lihao}
\end{figure*}

Being quadratic, the $\mathbf{X}$-subproblem can be readily solved by using the first-order optimality condition:
\begin{equation}
\label{equ:X_equations_WS_WD}
\begin{split}
&\left(\mu_1^{(k)}\mathbf V^{\top}\mathbf W_{\textrm{D}}^{\top}\mathbf W_{\textrm{D}}\mathbf V+\mu_2^{(k)}\mathbf B^{\top}\mathbf W_{\textrm{S}}^{\top}\mathbf W_{\textrm{S}}\mathbf B + \beta\sum_{i=1}^N {\mathbf S_i^T \mathbf S_i}\right)\mathbf X\\
&=\mathbf B^{\top}\mathbf W_{\textrm{S}}^{\top}\left( \mathbf Y_2^{(k)} +\mu_2^{(k)}\mathbf A^{(k+1)}\right) \\
&+ \mathbf V^{\top}\mathbf W_{\textrm{D}}^{\top} \left( \mathbf Y_1^{(k)} + \mu_1^{(k)} \left( \mathbf C^{(k+1)} + \mathbf W_{\textrm{D}}\tilde{\mathbf U}_f \right)\right) \\
&+ \beta\sum_{i=1}^N {\mathbf S_i^T \mathbf R_i^{(k+1)}}.
\end{split} \end{equation}
However, the straightforward matrix inversion in solving (\ref{equ:X_equations_WS_WD}) is inefficient or even practically impossible for large-scale problems, e.g., registration of tens of thousands of points. This can be relieved by using the LDL decomposition:
\begin{equation}
\begin{split}
\label{equ:LDL_decomposition_WS_WD}
&\left(\mathbf L, \mathbf D \right) = \\
&\textrm{ldl}\left( \mu_1^{(k)}\mathbf V^{\top}\mathbf W_{\textrm{D}}^{\top}\mathbf W_{\textrm{D}}\mathbf V+\mu_2^{(k)}\mathbf B^{\top}\mathbf W_{\textrm{S}}^{\top}\mathbf W_{\textrm{S}}\mathbf B + \beta\sum_{i=1}^N {\mathbf S_i^T \mathbf S_i}\right),
\end{split} \end{equation}
where $\mathbf L$ and $\mathbf D$ are the lower triangular matrix and the diagonal matrix of the LDL decomposition. Then, the linear equations in (\ref{equ:X_equations_WS_WD}) is solved by solving the following much easier linear systems:
\begin{equation}
\label{equ:X_solution_WS_WD}
\begin{split}
 \mathbf L \mathbf Q &= \mathbf V^{\top}\mathbf W_{\textrm{D}}^{\top} \left( \mathbf Y_1^{(k)} + \mu_1^{(k)} \left( \mathbf C^{(k+1)} +\mathbf W_{\textrm{D}} \tilde{\mathbf U}_f \right)\right) \\
 & + \mathbf B^{\top}\mathbf W_{\textrm{S}}^{\top}\left( \mathbf Y^{(k)} +\mu_2^{(k)}\mathbf A^{(k+1)}\right) + \beta\sum_{i=1}^N {\mathbf S_i^T \mathbf R_i^{(k+1)}} \\
\mathbf D \mathbf Z &= \mathbf Q,\\
\mathbf L^{\top} \mathbf X &= \mathbf Z.
\end{split} \end{equation}

The iterative non-rigid registration with reweighting is summarized in Algorithm 1, and the algorithm for minimization (\ref{equ:E_matrix_weighted}) is summarized in Algorithm 2.

 \begin{table}[!h]
\label{alg:NR_framework}
 \begin{center}
 \begin{tabular}{l}
  \hline
Algorithm 1. Algorithm of reweighting non-rigid registration \\
 \hline
    1. \quad Input: template $\mathcal V$, target $\mathcal U$.\\
    2. \quad \quad While not converged do\\
    3. \quad \quad \quad Find correspondence mapping $f^{(l)}:\mathcal V \mapsto \mathcal U $;\\
    4. \quad \quad \quad Update $\mathbf W^{(l)}_{\textrm{D}}$ and $\mathbf W^{(l)}_{\textrm{S}}$ acco. to (\ref{equ:weighting_dataterm}) and (\ref{equ:weighting_smoothterm}), resp.\\
    5. \quad \quad \quad Solve transformations $\mathbf X^{(l)}$ via Algorithm (2);\\
    6. \quad \quad End while\\
    7. \quad Output: $\mathbf X$\\
  \hline
 \end{tabular}
 \end{center}
\end{table}

\begin{table}[!h]
\label{alg:ADM_ALM}
 \begin{center}
 \begin{tabular}{l}
  \hline
Algorithm 2.  ADMM algorithm to solve (\ref{equ:E_matrix_weighted}) \\
 \hline
    ~1. \quad Input: $\tilde{\mathbf U}_{f^{(l)}}\in \mathbf R^{N\times 3}$, $\mathbf V\in \mathbf R^{N\times 4N}$, $\mathbf B\in \mathbf R^{|\mathcal{E}|\times 4|\mathcal{V}|}$;  \\
    ~2. \quad \quad Initialize: $\mathbf X^{(l,0)}=X^{(l-1)}$, $\mathbf Y_1^{(0)},\mathbf Y_2^{(0)}=0$; \\
    \quad \quad \quad \quad \quad \quad \quad \quad  $\mu_1, \mu_2>0$, $\rho_1, \rho_2>1$;\\
    ~3. \quad \quad While not converged do \\
    ~4. \quad \quad \quad  Solve $\mathbf C^{(l,k+1)}$ by (\ref{equ:C_solution_WS_WD});\\
    ~5. \quad \quad \quad  Solve $\mathbf A^{(l,k+1)}$ by (\ref{equ:A_solution_WS_WD});\\
    ~6. \quad \quad \quad  Solve $\mathbf R_i^{(l,k+1)}$ by (\ref{equ:R_solution_WS_WD});  \\
    ~7. \quad \quad \quad  Solve $\mathbf X^{(l,k+1)}$ by (\ref{equ:LDL_decomposition_WS_WD})$\sim$(\ref{equ:X_solution_WS_WD});  \\
    ~8. \quad \quad \quad  Update $\mu_1^{(k+1)}$, and $\mu_2^{(k+1)}$ according (\ref{equ:solveCAX_WS_WD});\\
    ~9. \quad \quad \quad  Update $Y_1^{(k+1)}$, and $Y_2^{(k+1)}$ according (\ref{equ:solveCAX_WS_WD});\\
    10. \quad \quad End while\\
    11. \quad Output: $\mathbf X^{(l)}$. \\
  \hline
 \end{tabular}
 \end{center}
\end{table}

\section{Experimental Results}
In this section, we evaluate the performances of the proposed method on clean datasets (Section \ref{sec:clean}), noisy datasets (Section \ref{sec:noisy}), and real scans (Section \ref{sec:real}). Running times of our method are reported in Section \ref{sec:time}.

\subsection{Results on Clean Datasets}
\label{sec:clean}

\begin{figure*}[ht]
  \centering
  \includegraphics[width=1.0\linewidth]{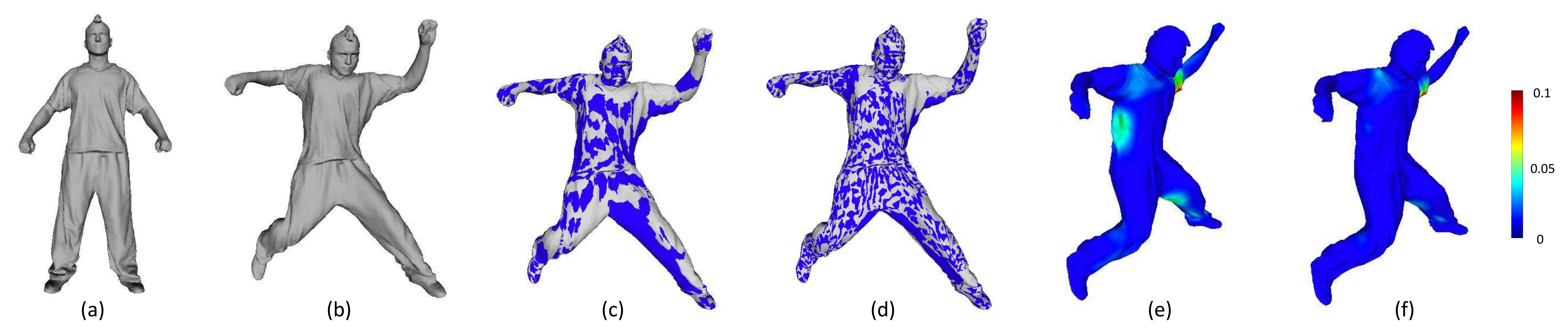}
  \caption{Comparison results of with and without reweighting scheme on \emph{Bouncing} dataset: (a) Template, (b) Target, (c) Registration result of without reweighting scheme, (d) Registration result of with reweighting scheme, (e) Fitting errors of without reweighting scheme, and (f) Fitting errors of with reweighting scheme.}
  \label{fig:reweight_clean}
\end{figure*}

\begin{figure*}[ht]
  \centering
  \includegraphics[width=0.95\linewidth]{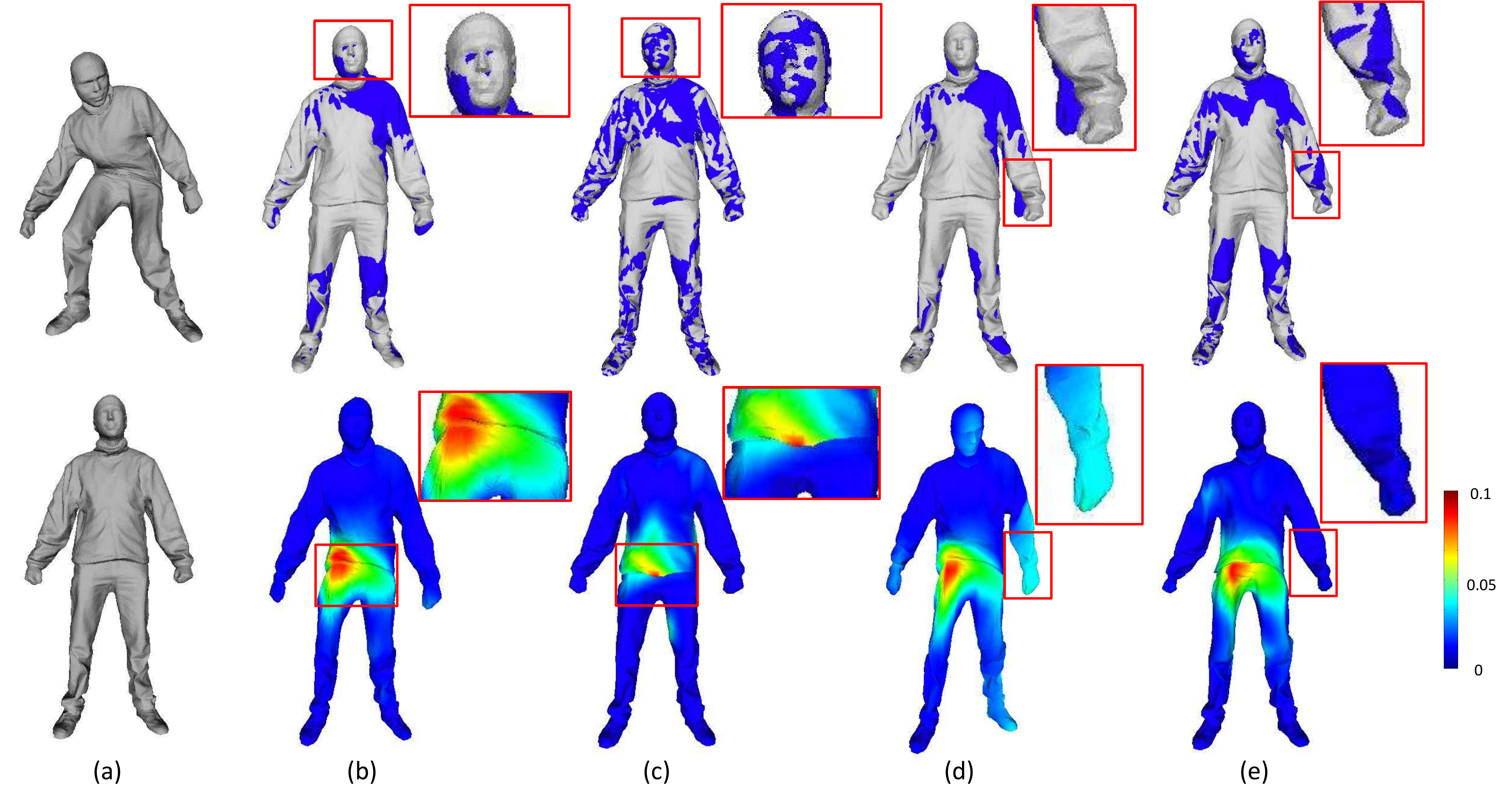}
  \caption{Comparison results on \emph{Jumping} dataset with partially incorrect correspondences: (a) Template and target,
(b) SNR method \cite{ySNR2015} result with one third SHOT correspondences, (c) Our method result with one third SHOT correspondences, (d) SNR method \cite{ySNR2015} result with all SHOT correspondences, and (e) Our method result with all SHOT correspondences.}
  \label{fig:shot}
\end{figure*}

We evaluate the proposed method on two datasets: TOSCA high-resolution dataset \cite{bronstein2008numerical} and a human motion dataset \cite{vlasic2008articulated}. Fig. \ref{fig:L2_L2L1_L1_cat} and Fig. \ref{fig:L2_L2L1_L1_human} give the registration results on \textit{cat} and \textit{jumping} datasets, compared with the classic $\ell_2$-norm regularized non-rigid ICP method  and the SNR method~\cite{ySNR2015}.
The results are shown as the overlap of the deformed template shape (blue) and the target shape (gray) and the fitting errors are color-coded on the reconstructed mesh for visual
inspection. Denote $\mathbf g_i$ as the ground-truth correspondence of $\mathbf v_i$. For a vertex $\mathbf v_i$, the registration error is defined as $ \|\mathbf X_i\mathbf v_i- \mathbf g_i \|^2_2$.
The compared classic $\ell_2$-norm based non-rigid ICP method is formulated as optimizing:
\begin{equation}\label{equ:E_ell_2_norm}
 \min_{\mathbf{X}}~~ \begin{Vmatrix}\mathbf W( \mathbf V\mathbf X-\tilde{\mathbf{U}}_f) \end{Vmatrix}_F^2 + \alpha \begin{Vmatrix}\mathbf B\mathbf X\end{Vmatrix}_F^2.
% ~~~\mathbf B \equiv \mathbf K\otimes \mathbf I_4
 \end{equation}
The smoothness constraint of this kind of methods is imposed on the transformation differences.
To ensure fair comparison, we adjust the weight $\alpha$ until we get the most accurate registration without loss of smoothness for each method. The result shows that our method achieves the best results with less fitting errors in the areas with intensive deformations than the SNR method \cite{ySNR2015} and the classic $\ell_2$-norm regularized non-rigid ICP method, such as the tail of
the cat and the wrinkles around the waist of the person highlighted in rectangles.

\begin{figure*}[ht]
  \centering
  \includegraphics[width=1.0\linewidth]{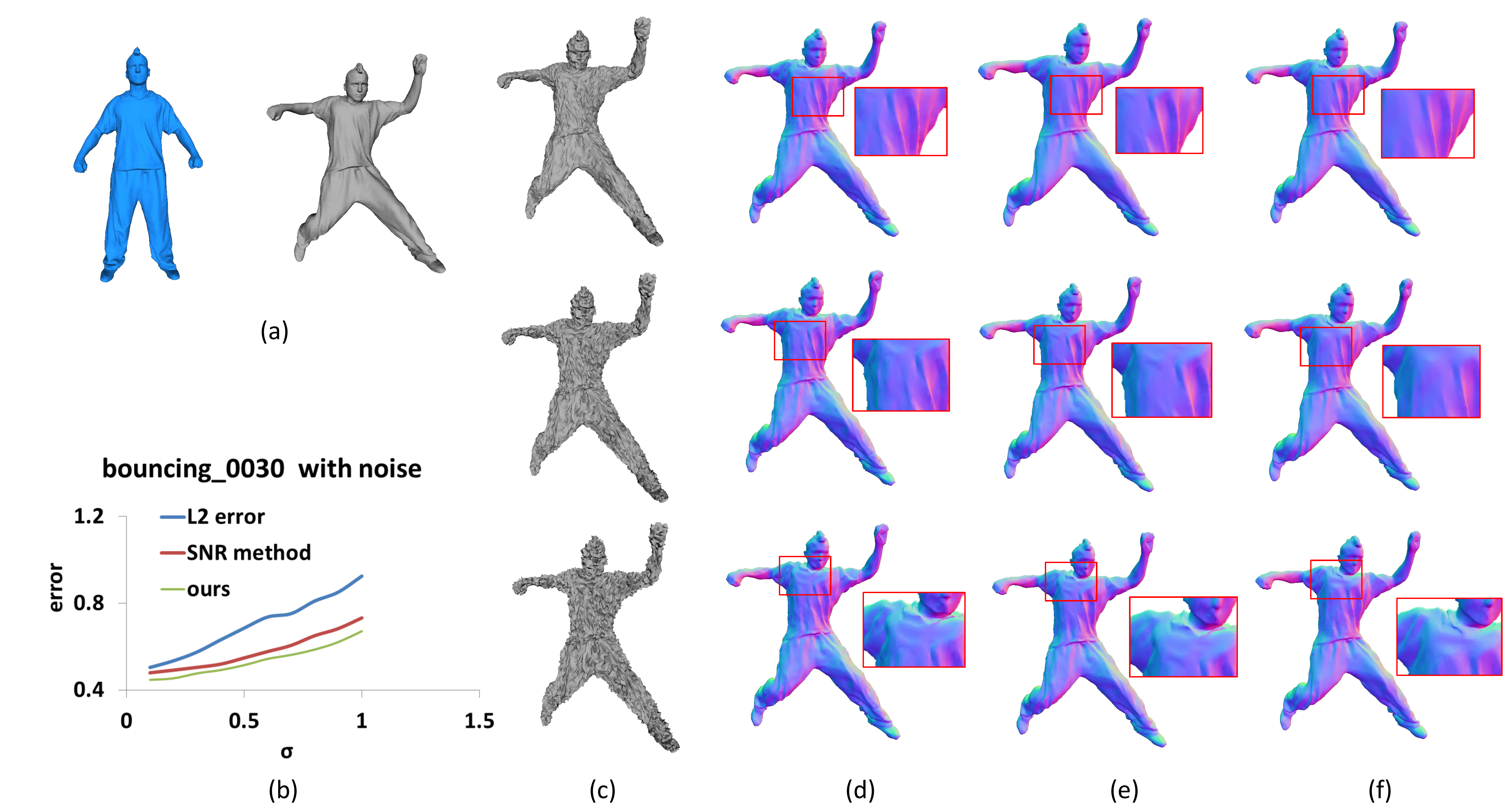}
  \caption{Comparison results on
\emph{Bouncing} with noise ($\sigma =0.3, 0.7, 1$). (a) Template and target, (b) Curves of fitting errors vs.
normalized noise levels, (c) Target with noise, (d) $\ell_2$-norm method, (e) SNR method \cite{ySNR2015}, and (f) Our method.  }
  \label{fig:noise}
\end{figure*}

\begin{figure*}[ht]
  \centering
  \includegraphics[width=1.0\linewidth]{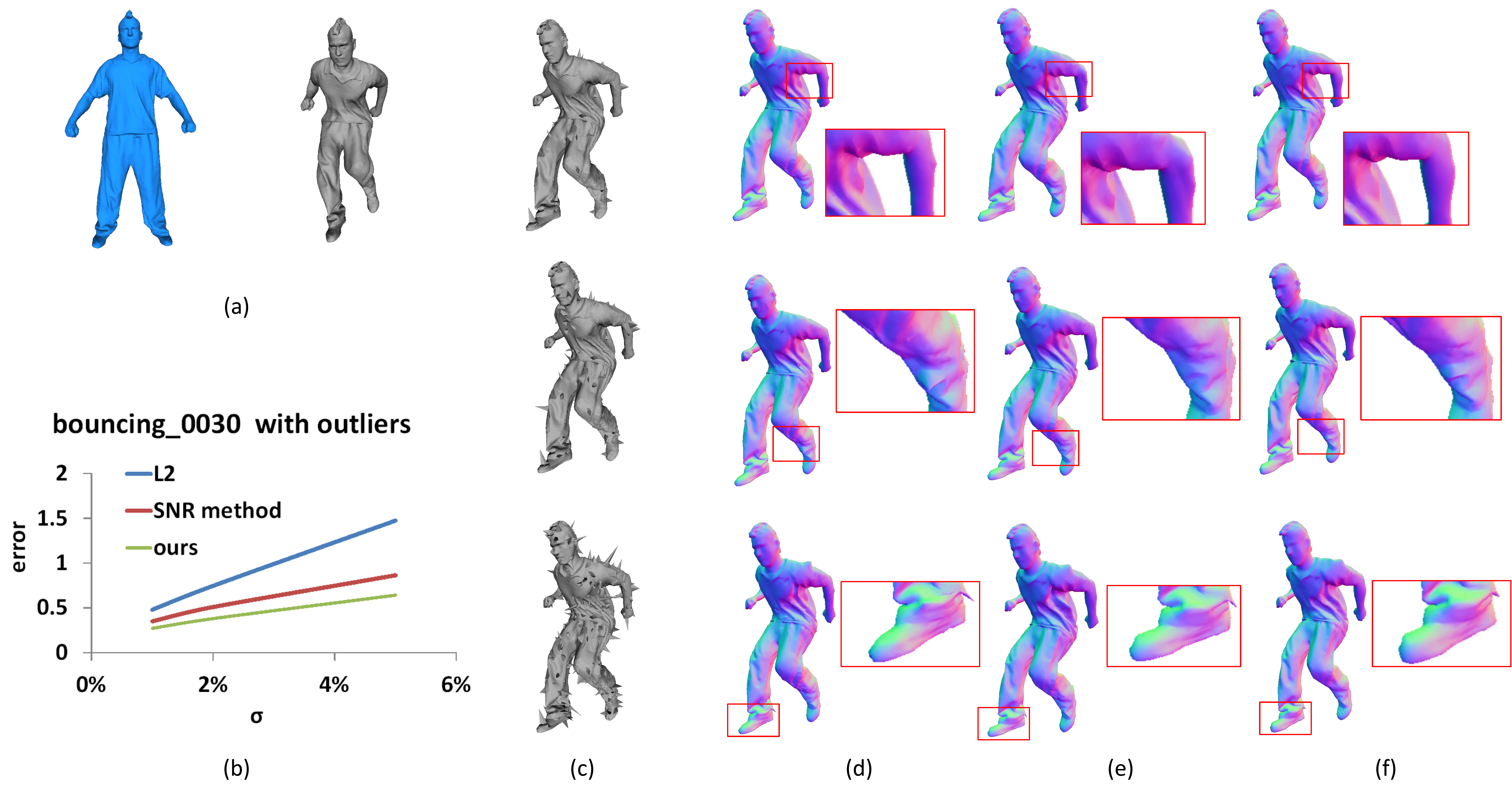}
  \caption{Comparison results on
\emph{Bouncing} with $1\%$, $2\%$,  $5\%$ outliers. (a) Template and target, (b) Curves of fitting errors vs.
normalized noise levels, (c) Target with noise, (d) $\ell_2$-norm method, (e) SNR method \cite{ySNR2015}, and (f) Our method.}
  \label{fig:outlier}
\end{figure*}

\begin{figure*}[ht]
  \centering
  \includegraphics[width=1.0\linewidth]{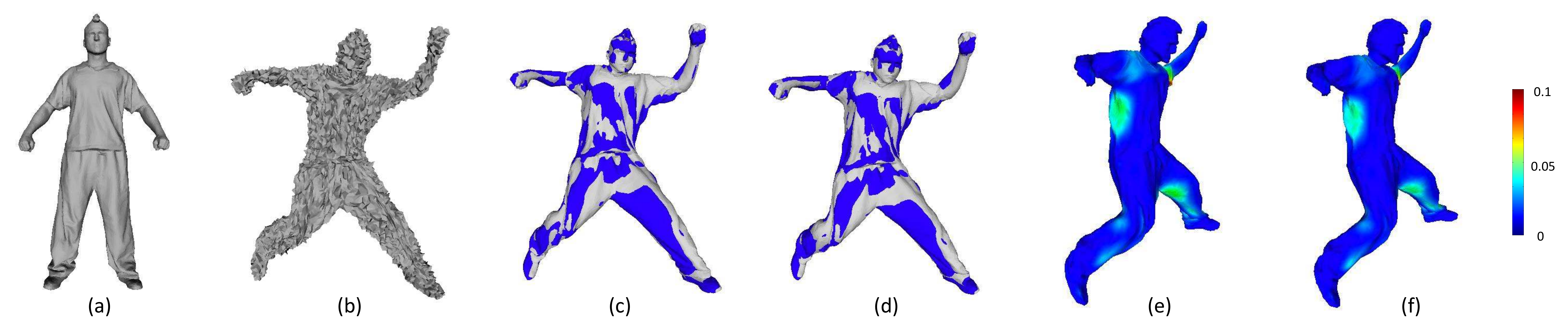}
  \caption{Comparison results of with and without reweighting scheme on \emph{Bouncing} dataset with noise ($\sigma =1$): (a) Template, (b) Target, (c) Registration result of without reweighting scheme, (d) Registration result of with reweighting scheme, (e) Fitting errors of without reweighting scheme, and (f) Fitting errors of with reweighting scheme.}
  \label{fig:reweight_noise}
\end{figure*}

\begin{figure*}[ht]
  \centering
  \includegraphics[width=1.0\linewidth]{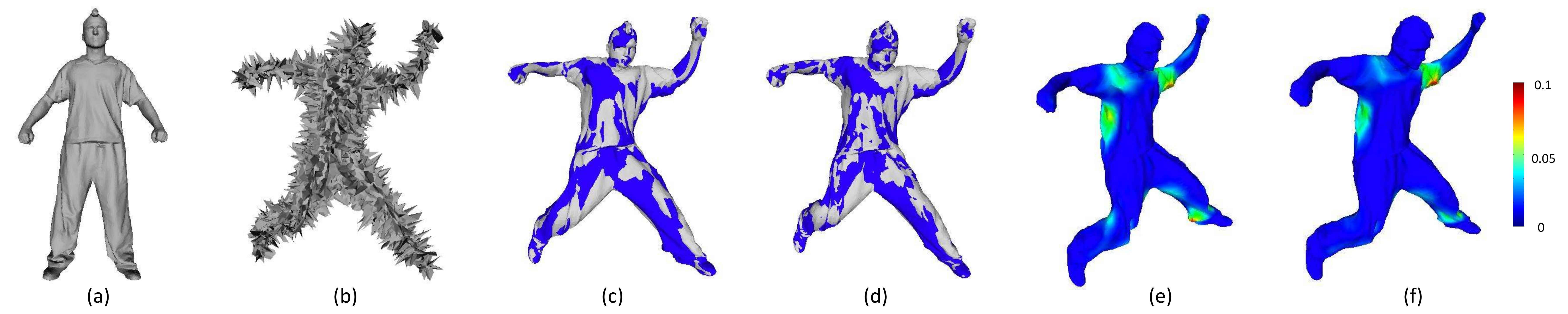}
  \caption{Comparison results of with and without reweighting scheme on \emph{Bouncing} dataset with $50\%$ outliers: (a) Template, (b) Target, (c) Registration result of without reweighting scheme, (d) Registration result of with reweighting scheme, (e) Fitting errors of without reweighting scheme, and (f) Fitting errors of with reweighting scheme.}
  \label{fig:reweight_outlier}
\end{figure*}

\begin{figure*}[ht]
  \centering
  \includegraphics[width=0.9\linewidth]{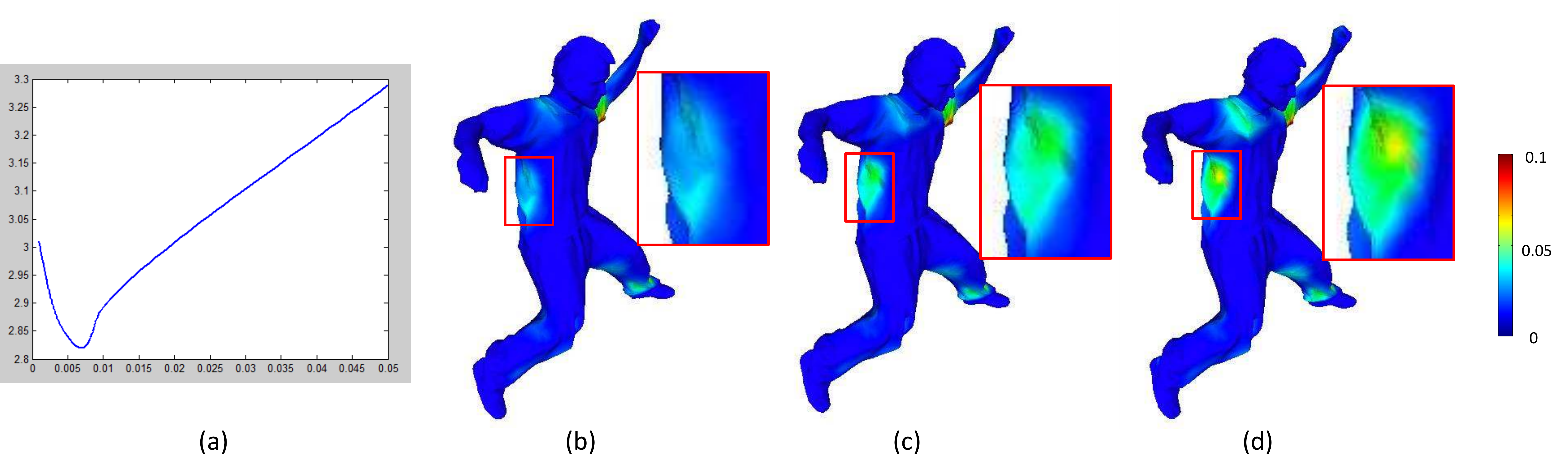}
  \caption{Comparison results with different parameter settings for the reweighting scheme on \emph{Bouncing} dataset with $50\%$ outliers: (a) Curves of fitting errors vs. $\epsilon$ values, (b) Registration result with $\epsilon = 0.006$, (b) Registration result with $\epsilon = 0.01$, and (d) Registration result with $\epsilon = 0.05$.}
  \label{fig:reweight_eps}
\end{figure*}

We also compare our method with state-of-the-art non-rigid registration \cite{li2008global} in Fig. \ref{fig:lihao}.  Obvious registration errors can be seen in the result of the method in \cite{li2008global}, especially in the right foot (top) and head (bottom), while the methods with sparse representation (SNR \cite{ySNR2015} and our method) achieve better registration results. The method in \cite{li2008global} works effectively when the template and target shapes are
close so that good initial correspondences can be obtained, but the pose changes substantially in this example. Moreover, our result is more accurate and better-distributed for the whole body than the SNR method \cite{ySNR2015}, due to the sparse constraint on the position.

To evaluate the robustness of the proposed method, we manually assign 35 correspondences on \emph{Jumping} dataset, and compare the result of our method with the SNR method \cite{ySNR2015} and the $\ell_2$-regularized method.
As shown in Fig. \ref{fig:few_corr}, our method achieves the best result, especially around the places with substantial deformation, e.g., the right knee.

To evaluate the effectiveness of the proposed reweighting scheme, we compare the registration results with and without reweighting on \emph{Bouncing} dataset in Fig. \ref{fig:reweight_clean}. The parameters $\epsilon_{\textrm{D}}$ and $\epsilon_{\textrm{S}}$ are set as 0.01. As shown in the figure, the reweighting scheme significantly improves the registration results.

\subsection{Results on Noisy Datasets}
\label{sec:noisy}

%This section evaluates the robustness of our method in the noisy cases.
\noindent\textbf{1) Correspondences with partially incorrect matchings:}

It is common to include incorrect correspondences using established methods. We simulate this in two cases. In the first case, we obtain two thirds of correspondences using diffusion pruning~\cite{tam2014diffusion} and the remaining one third using local geometric feature matching based on SHOT signatures~\cite{salti2014shot}.  The majority of correspondences from the former are correct while many correspondences from the latter are incorrect due to the ambiguity of local features. In the second case, we generate all the correspondences using SHOT signatures. Fig. \ref{fig:shot} gives the results for the two cases in a difficult situation which involves very complex transformations from template to target. As shown in the figure, our method is significantly more robust than the SNR method \cite{ySNR2015} with respect to incorrect correspondences.

\noindent\textbf{2) Target shapes with noise or outliers:}

In the first case, 3-D shapes of targets are polluted with dense noise along the norm directions of the associated vertices. All the target vertices are perturbed with Gaussian noise. The standard deviation of the noise $\sigma$ is normalized by $\bar{l}$, where $\bar{l}$ is the average length of triangle edges on the associated target mesh, and chosen in the range of $[0.1, 1]$.
Fig. \ref{fig:noise} gives the registration results compared with the SNR method \cite{ySNR2015} and the $\ell_2$-norm regularization method. The results show that our method is more robust to noise, performing significantly better for models with high noise levels.

In the second case, 3-D shapes of targets are polluted with sparse outliers along the normal directions of the associated vertices. Fig. \ref{fig:outlier} gives the results for the situations when $1\%$, $2\%$, $5\%$ of target vertices are perturbed with Gaussian noise. The results show that our method is more robust than the other two methods, particularly for cases with larger proportion of outliers.

To evaluate the effectiveness of the proposed reweighting scheme, we also compare the registration results with and without reweighting for noise and outlier cases on \emph{Bouncing} dataset in Fig. \ref{fig:reweight_noise} and Fig. \ref{fig:reweight_outlier}. The parameters $\epsilon_{\textrm{D}}$ and $\epsilon_{\textrm{S}}$ are set as 0.01. The standard deviation of the noise $\sigma$ is set as 1, and the percentage of outliers is set as $50\%$. It can be seen that the reweighting scheme contributes significantly to improve the registration results for the dataset with noise and outliers.

We compare the registration results with different parameter settings for the reweighting scheme on \emph{Bouncing} dataset with $50\%$ outliers in Fig. \ref{fig:reweight_eps} to evaluate the influence of the paremeters $\epsilon_{\textrm{D}}$ and $\epsilon_{\textrm{S}}$. To make experiments more tractable, we adjust both parameters consistently (i.e. $\epsilon_{\textrm{D}} = \epsilon_{\textrm{S}} = \epsilon$).  It can be seen that the best setting is 0.006 for this case, which has the smallest fitting errors. However, the performance is quite close, and 0.01 is a generally good choice (found in experiments).

\subsection{Results on Real Scans}
\label{sec:real}
\begin{figure*}[ht]
  \centering
  \includegraphics[width=6.0in]{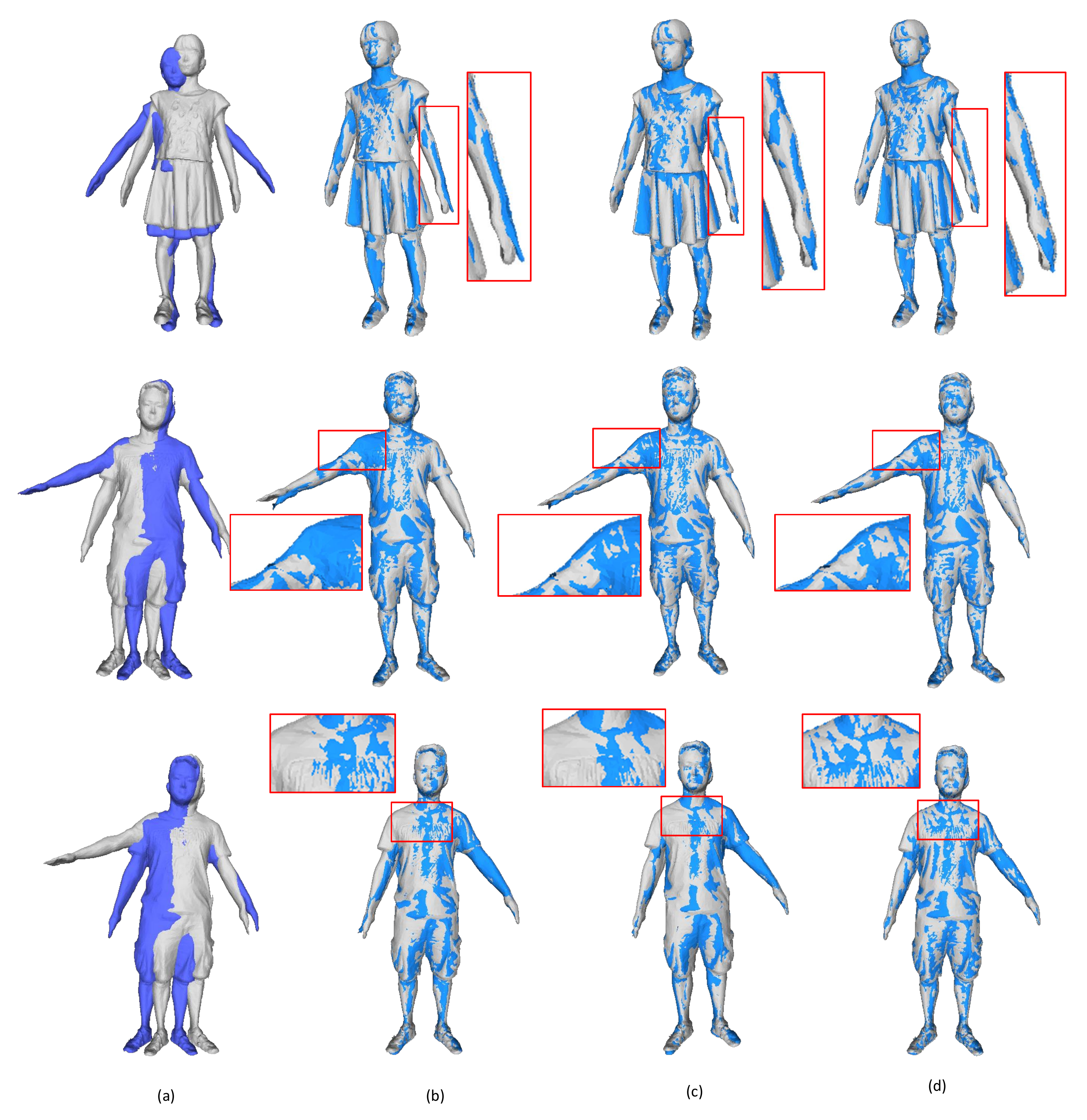}
  \caption{Comparison results on Kinect datasets: (a) Template
and target, (b) $\ell_2$-norm method, (c) SNR method \cite{ySNR2015}, and (d) Our method.}
  \label{fig:kinect}
\end{figure*}

Fig. \ref{fig:kinect} presents the results on real scans generated by Kinect Fusion \cite{kinectfusion} using Kinect V2.0.
The real scans are very challenging, because they have much noise and a large number of outliers. Moreover, each mesh is incomplete and the topology
between the template and the target is inconsistent. Hence, it is difficult to obtain sufficient and reliable correspondences.
The overlap of the deformed template and the target show that the $\ell_2$-norm regularization method and the SNR method present misalignments around the hands, arms and some other joints which have large deformations, while the result of our method is well-distributed and better registered.

\begin{figure*}[ht]
  \centering
  \includegraphics[width=0.9\linewidth]{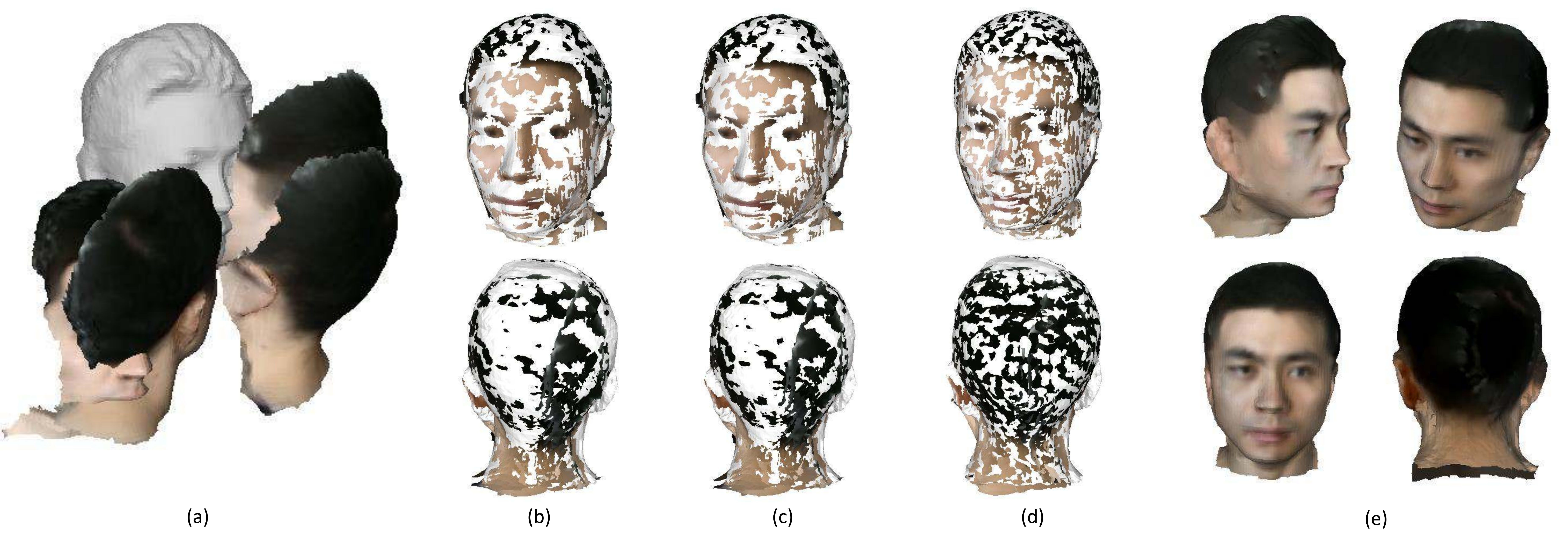}
  \caption{Comparison results on Kinect datasets: (a) Base mesh and four partial color meshes, (b) Registered results of $\ell_2$-norm method, (c) Registered results of SNR method \cite{ySNR2015}, (d) Registered results of our method, and (e) Texture fusion results of our method.}
  \label{fig:face}
\end{figure*}

Fig. \ref{fig:face} gives an example of generating a complete color mesh for a human head. A base mesh is scanned by Kinect Fusion using Kinect V2.0, and four partial color meshes are registered to the base mesh using our method. The textures are blended by solving the Poisson equation over the surface of mesh \cite{texturestitcher}. As shown in the figure, our method correctly registers
the input view surfaces with better registration than alternative methods, and successfully generates a watertight color mesh.

\subsection{Running times}
\label{sec:time}
We compare the running times of the proposed method with
the $\ell_2$-norm regularized method, SNR method, and group sparsity method on \emph{Crane} dataset.
We downsample the meshes into smaller meshes with 1K to 10K
vertices. The number of nICP registration iterations for each method is set as
20, and $\ell_1$-norm has extra 20 inner iterations for each outer iteration.
All the experiments are performed on a desktop computer
with Intel i5 3.2GHz CPU and 8GB RAM.
The comparison results are shown in Table 1. Our method has similar time complexity as SNR.

%\vspace{-.3cm}
\begin{table}[htbp]
\label{table1}
\centering
\caption{Comparison on running times}
\vspace{-.2cm}
\resizebox{0.45\textwidth}{!}{
\begin{tabular}{c|c|c|c|c}
\hline
Num. vertexes& 1000 &2000 &	5000 &	10000 \\
 \hline
$\ell_2$-norm&	1.23s&3.51s&12.88s&29.78s	\\
 \hline
SNR&8.05s&17.36s&52.48s&119.06s	\\
\hline
Group sparsity&7.39s&24.83s&59.96s&126.58s	\\
\hline
Ours&7.17s&22.13s&55.68s&122.85s	\\
\hline
\end{tabular}}
\end{table}

\vspace{-.1cm}
 \section{Conclusions}
This paper proposes a non-rigid registration  method with reweighted sparse position and transformation constraints.
We formulate the energy function with dual sparsity on both the data term and the smoothness term,
and define the smoothness constraint using local rigidity.
The dual-sparsity based non-rigid registration model is equipped with a reweighting scheme, and solved by
the alternating direction method under the augmented Lagrangian multiplier (ADM-ALM) framework which have exact solutions and guaranteed convergence.
Experimental results on both public datasets and real scans show that our method provides significantly improved results over alternative methods, especially for more challenging cases, and is more robust to noise and outliers.

\ifCLASSOPTIONcompsoc
  \section*{Acknowledgments}
\else
  \section*{Acknowledgment}
\fi

The authors would like to thank Ke Li for her help with some experiments, and thank Shuai Lin for help with comparative experiments with~\cite{li2008global}.

\bibliographystyle{ieee}
\bibliography{DS}

\end{document}